# A SELF-SUPERVISED TIBETAN-CHINESE VOCABULARY ALIGNMENT METHOD BASED ON ADVERSARIAL LEARNING

Enshuai Hou[1] and Jie zhu[2]

[1]Tibet University, Lhasa, Tibet, China
[1]hes@utibet.edu.cn
[2]Tibet University, Lhasa, Tibet, China
[2]790139756@qq.com

## ABSTRACT

*Tibetan is a low-resource language. In order to alleviate the shortage of parallel corpus between Tibetan and Chinese, this paper uses two monolingual corpora and a small number of seed dictionaries to learn the semi-supervised method with seed dictionaries and self-supervised adversarial training method through the similarity calculation of word clusters in different embedded spaces and puts forward an improved self-supervised adversarial learning method of Tibetan and Chinese monolingual data alignment only. The experimental results are as follows. First, the experimental results of Tibetan syllables Chinese characters are not good, which reflects the weak semantic correlation between Tibetan syllables and Chinese characters; second, the seed dictionary of semi-supervised method made before 10 predicted word accuracy of 66.5 (Tibetan - Chinese) and 74.8 (Chinese - Tibetan) results, to improve the self-supervision methods in both language directions have reached 53.5 accuracy.*

## 1. INTRODUCTION

Tibetan is a low-resource minority language, and the available Tibetan-Chinese sentences pair corpus is relatively scarce compared to English, Chinese, etc. However, research on tasks such as Tibetan-Chinese bilingual machine translation [1,2] requires a large number of bilingual comparisons. Compare Tibetan and Chinese bilingual corpus, large monolingual data more readily available. Extracting similar words which have semantic information from the Chinese and Tibetan monolingual corpus generates a comparison dictionary. This work can alleviate the need for bilingual comparison data in tasks such as machine translation. Based on Harris' [3] distribution hypothesis, words with similar contexts have similar semantics, and word vectors can reflect this distribution relationship to a large extent. The word vectors of similar words are relatively nearby to the embedding space and come from word clusters of different languages. Word clusters from different languages have similar distributions in different embedding spaces [4,5]. In this paper, semi-supervised, self-supervision and improved the self-supervision three kinds of Tibetan-Chinese bilingual glossary of methods, and the use of syllables and words two kinds of segmentation granularity experiment. In semi-supervised alignment method of mapping we use a seed dictionary to map, then spread throughout the semantic space; self- supervision alignment method, against networks [6] to learn mapping, by the mapping the representative word embedded mapping space to for aligned Tibetan Chinese vocabulary: The improved Tibetan-Chinese self- supervised vocabulary first uses a self-supervised method to learn to map, and then uses part of the high-frequency word pairs generated by this mapping as the seed dictionary, and iteratively improves the semi-supervised method to obtain the final Tibetan-Chinese aligned dictionary. The research of Chinese and Tibetan self-supervision vocabulary alignment method can effectively alleviate the need for bilingual data onto research and it has a positive meaning involving Chinese and Tibetan.

## 2. RELATED WORKS

There are many relevant pieces of research on obtaining cross-language vocabulary pairs from monolingual data onto different languages without using a grand number of parallel corpora. In 2013, Mikolov et al. [7] first used the hypothesis that words with different languages in similar contexts are similar to learn a linear mapping from the source of the target embedding space, using 5000 pairs of parallel words as anchor points for training, and evaluate their methods in a word translation task. The research of Xing et al. [8] in 2015 showed that compared with the method of Mikolov et al, the result can be improved by implementing orthogonal constraints on the linear mapping. In 2017, Smith et al. [9] used the same characters to construct a bilingual dictionary in an attempt to reduce the dependence on supervised bilingual dictionaries. In 2017, Artetxe et al. [10] initialized bilingual dictionaries with aligned numbers, and then gradually aligned the embedding space using an iterative method. However, their model can only be applied to languages with similar alphabetic languages due to the shared alphabet. Without using parallel corpus, Zhang et al. [11] began to try to use adversarial methods to obtain cross-lingual word embedding. Alexis et al. [12] used the adversarial training method to learn a linear mapping from the source language to the target language space, then used the Procrustes method of fine-tuning, and proposed a bilingual vocabulary similarity measurement method of cross-domain similarity local scaling. In Tibetan and Chinese language pairs, there is also the problem of a lack of data resources. Methods such as back translation [13] can expand the data and construct pseudo-parallel corpus, but the more demand for parallel sentence pairs and problems of quality cannot be avoided.

## 3. SELF-SUPERVISED TIBETAN-CHINESE VOCABULARY ALIGNMENT METHOD

This chapter mainly describes the Tibetan-Chinese vocabulary alignment method of the Tibetan-Chinese language differences, the Tibetan-Chinese vocabulary similarity measurement method, the semi-supervised method with a seed dictionary, the self-supervised method, and the improved self-supervised method. According to the research content of this article, firstly, it introduces the differences between Tibetan and Chinese languages from the question of whether the constituent elements of Tibetan and Chinese languages require segmenting. Secondly, it introduces the method of measuring similarity between Tibetan and Chinese vocabularies and their use. Then, from the perspective of the model, the semi-supervised seed dictionary method, the self-supervised Tibetan-Chinese vocabulary alignment method based on adversarial training, and the improved method based on self-supervised alignment with the semi-supervised method is described.

### 3.1. Differences Between Tibetan and Chinese

Tibetan is a phonetic script and has its own special language organization. The smallest morpheme in Tibetan is Tibetan syllable, and the smallest semantic unit is Tibetan words. A Tibetan syllable is composed of one or more Tibetan characters, and Tibetan words can be composed of one or more syllables, but there is no obvious division of words, and there is still a phenomenon of deflation [14]. The representation of the same thing in different dialects is also different, causing many processing difficulties and increasing the difficulty of labeling Tibetan data.

Chinese is a square text, the semantic unit is a word, one or more Chinese characters can form a Chinese vocabulary, there are no separators between words, and there are many polysemous phenomena in a word. Comparing Tibetan and Chinese, two scripts with the same semantic length, in a computer, Tibetan requires more storage space, etc.

## 3.2. Similarity Measurement Method of Tibetan and Chinese Vocabulary

In order to measure the similarity between Tibetan and Chinese words, this paper uses Cross-domain Similarity Local Scaling(CSLS) as a measure of similarity between Tibetan and Chinese words. This method is based on the improvement of the K-nearest neighbor method, and solves the problem of the K-nearest neighbor method of high-dimensional space: the nearest neighbor is asymmetric. In the high-dimensional space, some words are the nearest neighbors of many words, and some words are not neighbors of any words. The nearest neighbor and the average similarity of the nearest neighbor is introduced as a penalty factor of the CSLS method, which improves the local accuracy.

The CSLS method is defined as follows: Suppose there is a two-part neighborhood graph in the vector space of two languages, where each word is connected to $K$ nearest neighbors in the other language. Then respectively calculate the cosine similarity between the word and the $K$ nearest neighbors, and finally take the average value to obtain the average similarity $r$ between a vocabulary and adjacent vocabulary in another language.

$$r_T(Wx_s) = \frac{1}{K}\sum_{y_t \in \mathcal{N}_T(Wx_s)} \cos(Wx_s, y_t) \quad (1)$$

Combining the two conversion directions of the source language to the target language and the target language to the source language, the cosine similarity is combined with r in the two directions as the penalty factor to form the similarity measurement method CSLS:

$$\text{CSLS}(Wx_s, y_t) = 2\cos(Wx_s, y_t) - r_T(Wx_s) - r_S(y_t) \quad (2)$$

This paper will use this similarity measurement method in two aspects: one is to use it in the process of generating bilingual alignment dictionaries; the other is to use it in the result evaluation index, taking the accuracy of the first N similar words, that is, P@N as the evaluation index.

## 3.3. Semi-supervised Method with Seed Dictionary

For Tibetan and Chinese, use the monolingual corpus to train word vectors, and obtain two sets of vectors with dictionary size n, m. Source language vector is defined as: $\{x_1, ..., x_n\}$, and Target language vector is defined as: $\{y_1, ..., y_m\}$.

Based on these two sets of vectors, a cross-language vocabulary alignment method is learned. First, use the word vectors generated by the two monolingual corpora of Tibetan and Chinese, and use some word pairs $\{x_i, y_i\}_{i \in \{1,n\}}$, $x \in X, y \in Y$ in the seed dictionary, as n pairs anchor point, by minimizing $\|WX - Y\|$, learns the mapping $W$, as shown in formula 3.

$$W^* = \underset{W \in M_d(\mathbb{R})}{\arg\min} \|WX - Y\| \quad (3)$$

Where $d$ is the dimension of the word vector, and $W$ belongs to a $d \times d$ real matrix $M_d(\mathbb{R})$ X and $Y$ are $d \times n$ dictionary embedding matrices to be aligned. Secondly, according to the research of Xing et al. [8], orthogonal constraints are implemented on $W$ to improve the results, and $W$ is an orthogonal matrix through singular value decomposition constraints, as shown in formula 4. Among it d is the dimension of the word vector, and $W$ belongs to a matrix of real numbers. And are to be aligned with the size of the word typical embedded matrix.

$$W^* = \underset{W \in O_d(\mathbb{R})}{\arg\min} \|WX - Y\|_F = UV^T, U\Sigma V^T = SVD(XY^T) \quad (4)$$

Through iterative training in two alignment directions, update $W$, minimize the difference between $WX$ and $Y$, and use mapping $W$ to align vocabulary to generate Tibetan-Chinese bilingual word pairs $\{Wx_i, y_i\}_{i\in\{1,t\}(t<m,t<n)}$, changes the translation direction and repeat the experiment, use the following method to generate two translation direction dictionaries.

In the dictionary generation process, on $\{Wx_1, ..., Wx_n\}$ and $\{y_1, ..., y_m\}$, uses the CSLS method to calculate the CSLS value of K neighbors in two directions. Based on the seed dictionary, the distance between the two languages dictionary is updated for the closest word pair, and a new aligned dictionary is finally generated.

### 3.4. Self-supervision Methods

The self-supervised method uses a generative adversarial network, defines the discriminator as a fully connected network, and the generator is a randomly initialized linear mapping $W$, and learns the latent space between Tibetan and Chinese through the training of the discriminator and generator. The discriminator is used to distinguish the elements from $WX = \{Wx_1, ..., Wx_n\}$ and $Y$. The role of the discriminator is to distinguish the two embedding sources. The mapping function $W$ to be learned by the generator makes it difficult for the discriminator to distinguish whether the word embedding is $Wx$ $(x \in X)$ or $y$ $(y \in Y)$..

The parameter that defines the discriminator network is $\theta_D$. When we assume z is a word embedding vector of unknown source, $P_{\theta_D}(source = 1|z)$ represents the probability that the discriminator considers the vector z to be the source language embedding $P_{\theta_D}(source = 0|z)$ means that the discriminator considers the probability that the vector z is embedded in the target language. Here, the cross-entropy loss is used as the loss of the discriminator. The formula of loss function is as follows:

$$L_D(\theta_D \mid W) = -\frac{1}{n}\sum_{i=1}^{n} \log P_{\theta_D}(\text{source } = 1 \mid Wx_i) - \frac{1}{m}\sum_{i=1}^{m} \log P_{\theta_D}(\text{source } = 0 \mid y_i) \quad (5)$$

Correspondingly in self-supervised training, for generator mapping $W$, its loss is defined as follows:

$$L_W(W \mid \theta_D) = -\frac{1}{n}\sum_{i=1}^{n} \log P_{\theta_D}(\text{source } = 0 \mid Wx_i) - \frac{1}{m}\sum_{i=1}^{m} \log P_{\theta_D}(source = 1 \mid y_i) \quad (6)$$

This article conducts the training of the adversarial network according to the standard adversarial network training process described by Goodfellow et al. [6] For a given input sample, the discriminator and the mapping matrix $W$ are updated using the stochastic gradient descent method to minimize $L_D$ and $L_W$ and finally make the two loss functions no longer drop. At the same time, the orthogonal constraint is added when updating $W$ during the training process, and it is used alternately with gradient descent. The constraint formula is shown in formula 7. Orthogonal constraints can maintain the dot product of the vector, ensure the quality of the word embedding corresponding to the language, and make the training process more robust. The value of $\beta$ is generally 0.001 to ensure that $W$ can almost always remain orthogonal.

$$W \leftarrow (1 + \beta)W - \beta(WW^T)W \quad (7)$$

Finally, a self-supervised Tibetan-Chinese bilingual aligned dictionary is generated using the dictionary generation method in the semi-supervised method with seed dictionary mentioned before.

### 3.5. Improve Self-supervised Adversarial Learning Method

In the self-supervised adversarial method, the mapping relationship of high-frequency words can better reflect the global mapping relationship. From the self-supervised adversarial method generation dictionary, select *s* words pairs with a higher frequency of occurrence $\{Wx_i, y_i\}_{i \in \{1,s\}}$ as anchor points, and then use the semi-supervised method to improve Training, iteratively updates the mapping $W$, extend the mapping to the lower frequency vocabulary domain, and generate a new alignment dictionary. The dictionary generation process is the same as the semi-supervised method of the seed dictionary.

## 4. EXPERIMENT AND ANALYSIS

### 4.1. Data Collection

Two separate corpora of Tibetan and Chinese were collected from the Internet, each with more than 35,000 sentences. The two-grained segmentation processing was performed on Tibetan and Chinese respectively, and experiments on the syllable size and word size of both Tibetan and Chinese characters were carried out.

The two dictionary pairs, Tibetan-Chinese and Chinese-Tibetan, are constructed artificially as the anchor point and test dictionary of the semi-supervised training method. The word granularity dictionary size is 10000, and the syllable granularity dictionary size is 3000.

### 4.2. Parameters Settings

The word vector training adopts the skip-gram model of the fasttext model, and the vector dimension is 300. The training framework uses Pytorch, and the specific parameter settings are shown in Table 1 and Table 2. The semi-supervised method and the improved part of the improved self-supervised method use consistent parameters.

Table 1 Semi-supervised and improved self-supervised parameter Settings

| parameter | value |
|---|---|
| **Seed dictionary size (syllables)** | 1500 |
| **Seed dictionary size (words)** | 5000 |
| **Test set size (syllables)** | 500 |
| **Test set size (words)** | 1500 |
| **Number of iterations** | 5 |
| **Word vector normalization processing** | True |
| **Generate a dictionary word on the number of** | 50000 |

Table 2 Self-supervised adversarial training parameter setting

| parameter | value | parameter | value |
|---|---|---|---|
| **Test set size (syllables)** | 500 | Number of training cycles per training | 100000 |
| **Test set size (words)** | 1500 | Discriminator network layer number | 2 |
| **Word vector normalization processing** | True | Number of single-layer nodes | 2048 |
| **Generate dictionary word pairs** | 50000 | Activation function | LeakReLU |
| **Training times** | 5 | Network input discard rate | 0.1 |
| **Orthogonalization parameters** | 0.001 | Training batch size | 32 |

### 4.3. Results analysis

This paper conducts two methods of experiments on Tibetan and Chinese monolingual corpora with syllable granularity and word granularity in two alignment directions. One is the accuracy of the word granularity experiment under (1, 5, 10) candidate words, which is P@N. Experiment, and compare the experimental effects of Tibetan → Chinese (Ti-Zh) and Chinese → Tibet (Zh-Ti); the second is the accuracy of the word granularity under (1, 5, 10) candidate words, P@N's experiment, and compared the experimental effects of Tibet→Chinese (Ti-Zh) and Chinese → Tibet (Zh-Ti). In tables, these use Semi-sup, Self-sup, Self-sup-re to Means semi-supervised method, self-supervised adversarial and the refine method of self-supervised adversarial.

In the syllable granularity experiment, as shown in Table 3, as the number of candidate words increases, the accuracy of the semi-supervised model P@N gradually increases, but the overall is bad although the self-supervised method has an increasing trend, In two directions it is difficult to learn the semantics of embedding space in this direction; after the improvement, the effect cannot be improved, and even the performance has declined in several results, showing a kind of disorder, which proves the weak semantic connection between Tibetan syllables and Chinese characters.

Table 3 Syllable granularity the value of P@N in both alignment directions

|  | Ti-Zh | | | Zh-Ti | | |
| --- | --- | --- | --- | --- | --- | --- |
|  | P@1 | P@5 | P@10 | P@1 | P@5 | P@10 |
| **Semi-sup** | 11.8 | 24.8 | 30.9 | 14.4 | 24.2 | 28.6 |
| **Self-sup** | 0.2 | 1.0 | 1.0 | 0.2 | 0.4 | 1.0 |
| **Self-sup-re** | 0 | 1.0 | 1.6 | 0 | 0.2 | 0.6 |

In terms of word granularity experiments, as shown in Table 4, compared to the self-supervised method, the semi-supervised method has achieved good results, and the improved Tibetan-Chinese self-supervised alignment method has achieved good results in the direction of Tibetan →Chinese. Both reached 53.5. In this process, the improved training played a great role and significantly improved the experimental effect. Although the improved self-supervised method has achieved a weaker effect than the semi-supervised method, it is of positive significance because it does not use the contrast dictionary for training.

Table 4 P@N values in the two alignment directions under the granularity of Tibetan and Chinese words

|  | Ti-Zh | | | Zh-Ti | | |
| --- | --- | --- | --- | --- | --- | --- |
|  | P@1 | P@5 | P@10 | P@1 | P@5 | P@10 |
| **Semi-sup** | 48.5 | 62.7 | 66.5 | 55.7 | 69.8 | 74.8 |
| **Self-sup** | 12.7 | 23.7 | 29.2 | 8.4 | 16.8 | 21.7 |
| **Selfs-up-re** | 25.4 | 46.3 | 53.5 | 27.5 | 47.7 | 53.5 |

After training, the word vectors corresponding to the partially aligned vocabulary of the two languages of Tibetan and Chinese are subjected to principal component analysis (PCA), and the two-dimensional vectors are visualized on the plane. The result is shown in figure 1. It can be seen from the figure that Tibetan and Chinese words have similar meanings in similar positions in space.

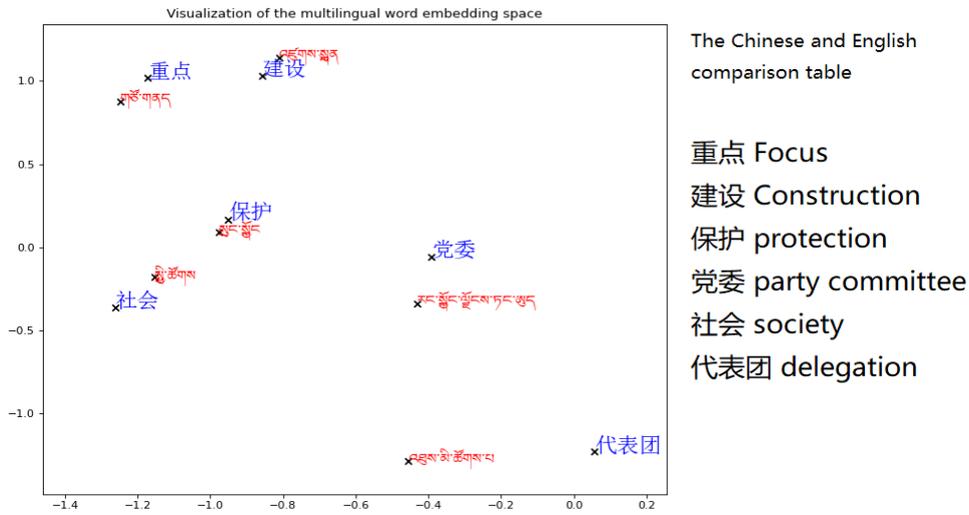

Fig.1 PCA visualization of partial results

## 5. CONCLUSIONS

In order to alleviate the need for parallel corpus for bilingual sentences in tasks such as Tibetan-Chinese translation, this paper uses the assumption that words in similar contexts are similar in different languages, using semi-supervised and self-supervised methods from Tibetan-Chinese monolingual data Extract aligned Tibetan-Chinese bilingual vocabulary and construct a bilingual aligned dictionary. First, the experiment of syllable granularity proves the weak correlation between Tibetan syllables and the semantics of Chinese characters. Second, improved self-supervision methods have achieved relatively effective results in the experiment of word granularity. The research in this article can alleviate the scarcity of Tibetan-Chinese parallel corpus data and provide a good start for unsupervised Tibetan-Chinese machine translation.

## ACKNOWLEDGEMENTS

Foundation Project: "Construction of Tibetan and Chinese Bidirectional Machine Translation Platform for the Inheritance and Development of Tibetan Language" and " Tibetan language training technology and Application".

**Authors**

Enshuai Hou 1996.09 Postgraduate in Tibet University Lhasa Tibet China

Direction: Tibetan-chinese machine translation.

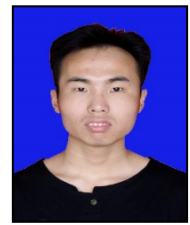

Jie Zhu 1973.11 Professor doctoral supervisor in Tibet University Lhasa Tibet China

Direction: Natural language processing (NLP), Tibetan information processing,

Data Mining and artificial intelligence.

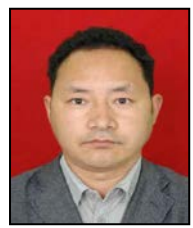